# Belief Revision in Probability Theory

Pei Wang
Center for Research on Concepts and Cognition
Indiana University
Bloomington, IN 47408
*pwang@cogsci.indiana.edu*

## Abstract

In a probability-based reasoning system, Bayes' theorem and its variations are often used to revise the system's beliefs. However, if the explicit conditions and the implicit conditions of probability assignments are properly distinguished, it follows that Bayes' theorem is not a generally applicable revision rule. Upon properly distinguishing belief revision from belief updating, we see that Jeffrey's rule and its variations are not revision rules, either. Without these distinctions, the limitation of the Bayesian approach is often ignored or underestimated. Revision, in its general form, cannot be done in the Bayesian approach, because a probability distribution function alone does not contain the information needed by the operation.

## 1 INTRODUCTION

In a reasoning system that deals with uncertainty, a proposition can be represented as $A[m]$, where $A$ is a sentence of a formal language, and $m$ indicates the sentence's uncertainty.

In different systems, $A$ and $m$ may have different forms and interpretations, and the operations on them may be defined differently. However, there are still operations shared by many systems, in spite of all the differences (Bhatnagar and Kanal 1986):

**Comparison:** To decide which of the $A_i[m_i]$ has the highest certainty by comparing $m_i$ ($i = 1, 2, \cdots, n$).

**Propagation:** To get a conclusion $A_{n+1}[m_{n+1}]$ from a set of premises $A_i[m_i]$, where $A_{n+1}$ is different from $A_i$ ($i = 1, 2, \cdots, n$).

**Revision:** To modify the uncertainty value of a proposition $A$ from $m$ to $m'$ in the light of other propositions.

Defined as above, propagation (or inference) and revision (or combination) are clearly different.

In propagation, the system *generates a new proposition* (with its certainty value) that was not among the premises.

In revision, the system *modifies the certainty value of an existing proposition*.

Other authors may use the two words differently (Pearl 1988), but we can still make the above distinction, no matter how the operations are named.

Bayes' theorem seems to be an exception. In some systems, it is used as a propagation rule, while in others, as a revision rule. Let's begin our analysis with these two usages, which will lead us to the kernel of the debate on the limitations of the Bayesian approach as a model of reasoning with uncertainty.

## 2 PROPAGATION VS. REVISION

According to probability theory, it is necessary to define a proposition space $S$ in order for a system to represent and calculate probabilities of propositions. $S$ is the set of all propositions to be processed by the system, and may be generated from a set of atomic propositions, using logical operators (Wise and Henrion 1986).

As the starting point of all probability calculations, a prior probability distribution should be defined on $S$, under the constraints of the axioms of probability theory.

To choose such a prior probability distribution for a specific problem domain, some background knowledge (such as statistical data or subjective estimates) and general principles (such as the principle of indifference or maximum entropy) are necessary. Let's refer to them collectly as $C$, the *implicit condition* of the distribution function, and write the prior probability distribution function as

$$P_C : S \to [0, 1].$$

From Bayes' theorem, we can get the conditional probability of $A_1$ under the condition that $A_2$ is true (obviously, the conditional probability is also based on $C$):

$$P_C(A_1|A_2) = \frac{P_C(A_1 \wedge A_2)}{P_C(A_2)} \quad (1)$$

where $A_1$ and $A_2$ are both in $S$ (so $A_1 \wedge A_2$ is in $S$, too), and $P_C(A_2) > 0$. To prevent confusion, I call $A_2$ the *explicit condition* of the conditional probability assignment, to distinguish it from $C$.

Considering the three previously defined operations, we can see that Bayes' theorem, when used like this, is actually a propagation rule, not a revision rule, for the following reasons:

1. The conclusion is a *conditional* proposition, differing from the premises, which are both *unconditional*.

2. The rule derives a *new* probability assignment, instead of modifying a previous assignment. The prior (unconditional) probability assignment of $A_1$ is still valid and available for future usages.

3. It is unnecessary for the system to know whether $A_2$ is really true when applying the theorem. The conditional probability is gotten under the *assumption* that $A_2$ is true. If the system has enough resources, it can calculate all possible conditional probabilities using only the prior probabilities, without any "new information" that has not been included in $C$.

Used in this way, Bayes' theorem is a rule for propagating probability assignments from prior probabilities $P_C(x)$ to conditional probabilities $P_C(x|y)$, not a revision rule for changing $P_C(x)$ to a new probability distribution $P_{C'}(x)$, according to new evidence.

But when we perform the previously mentioned "comparison" operation, usually what we are concerned with is the probability of propositions *under all available evidence*.

For example, If we want to know which of $A_1$ and $A_2$ is more likely to be true, we'll compare $P_C(A_1)$ and $P_C(A_2)$ at the very beginning, when all of our knowledge about the problem is $C$. Later, if we get new evidence that shows $A_3$ to be true ($A_3 \in S$), we will compare $P_C(A_1|A_3)$ and $P_C(A_2|A_3)$, which can be calculated according to (1). In such a case, to compare the "probability" of $A_1$ and $A_2$, the implicit condition $C$ and the explicit condition $A_3$ are merged together to become "all available evidence".

Since the distinction between these two types of conditions no longer seems necessary here, it is possible to "compile" all of the explicit conditions that turn out to be true into the implicit condition of the probability distribution function, transforming the related conditional probabilities into new unconditional probabilities. This is the "conditionalization principle" (Earman 1992, Levi 1983, and Weirich 1983).

To distinguish this usage from the previous one of Bayes' theorem, a new function $BEL(x)$ can be defined on $S$ (Pearl 1986), which representing the probability distribution under all available evidence:

$$BEL(x) = P_C(x|K) \qquad (2)$$

where $x \in S$, and $K$ is the current evidence, that is, the conjunction of all propositions in $S$ that are known to be true.

Similarly, we can define a conditional distribution for $BEL(x)$ as follows (when $P_C(y|K) > 0$):

$$BEL(x|y) = \frac{P_C(x \wedge y|K)}{P_C(y|K)} \qquad (3)$$

Consistently with previous definitions, we refer to $y$ as the probability assignment's *explicit condition*, and to $C$ and $K$ as its *implicit condition*.

What makes $BEL(x)$ different from $P(x)$ is: "all evidence" is a dynamic concept for an open system which constantly accepts new knowledge from its environment, so $BEL(x)$ is time-dependent.

Let's use $BEL_t(x)$ to indicate its values at time $t$ (here time is measured discretely by counting the coming evidence).

At time 0 (the initial state), all given knowledge is in $C$, so

$$BEL_0(x) = P_C(x) \qquad (4)$$

Assuming at time $t$ the current evidence is $K_t$. If new evidence shows that $A$ is true ($A \in S$, and $P_C(A|K_t) > 0$), then, at time $t + 1$, $BEL(x)$ becomes

$$\begin{aligned} BEL_{t+1}(x) &= P_C(x|K_{t+1}) \\ &= P_C(x|A \wedge K_t) \\ &= \frac{P_C(x \wedge A|K_t)}{P_C(A|K_t)} \end{aligned} \qquad (5)$$

From (2), (3) and (5), we get

$$\begin{aligned} BEL_{t+1}(x) &= BEL_t(x|A) \\ &= \frac{BEL_t(x \wedge A)}{BEL_t(A)} \end{aligned} \qquad (6)$$

Comparing (6) to (1), we can see that unlike $P(x)$'s distribution, $BEL(x)$'s distribution is *modified* from time to time by applying Bayes' theorem to transform true explicit conditions into the implicit condition of the distribution function. To $BEL(x)$, Bayes' theorem is indeed a revision rule, not a propagation rule.

Under the assumption that each piece of evidence is a proposition represented as $A[m]$ (according to the convention at the beginning of the paper), we can easily list the preconditions for using Bayes' theorem as a revision rule for a probability distribution which representing the system's current beliefs, considering all available evidence and background knowledge implicitly:

1. $m \in \{0, 1\}$, that is, the new evidence is binary-valued, so it can be simply written as $A$ or $\neg A$.

2. $A \in S$, otherwise its probability is undefined.

3. $P_C(A) > 0$, otherwise it cannot be used as a denominator in Bayes' theorem.

## 3 EXPLICIT CONDITION VS. IMPLICIT CONDITION

Why do we need a revision rule in a plausible reasoning system?

We are interested in the truth values of a set of propositions $S$, but our knowledge about them is incomplete or inaccurate. At the very beginning, we have some background knowledge $C$, which provides the prior probability distribution for the system. Later, when the system get new knowledge $C'$, we want it to adjust its probability distribution to summarize both $C$ and $C'$. In such a way, the system can learn from its experience, and the defects in $C$ are remediable.

Of course, every information processing system has restrictions about the type of new knowledge that can be accepted. However, it is reasonable to expect that the domain knowledge which can be put into the system *a priori* (in $C$), can also be put into it *a posteriori* (in $C'$).

Now we can see why I distinguish implicit conditions from explicit conditions: for the Bayesian approach, an implicit condition is the knowledge that we can put into a probability distribution initially, and an explicit condition is the knowledge that the system can learn hereafter by using Bayes' theorem as a revision rule.

Therefore, our original question about whether Bayes' theorem can be used as a generally applicable revision rule can be broken down into three questions:

1. What type of knowledge can be represented as an explicit condition?
2. What type of knowledge can be represented as an implicit condition?
3. What is the relation between them?

The first question is answered by the three preconditions given at the end of the previous section. I claim that these preconditions cannot be applied to implicit conditions in general, for the following reasons:

1. An explicit condition must be a binary proposition, but an implicit condition can include statistical conclusions and subjective probabilistic estimates.
2. An explicit condition must be in $S$, but knowledge in an implicit condition only need to be related to $S$. For example, "Tweety is a bird and cannot fly" can be part of an implicit condition, even though $S$ includes only "Birds can fly", and does not include the name "Tweety" at all.
3. If a proposition is assigned a prior probability of zero according to $C$, it cannot be used as an explicit condition to revise the function. However, in practical domains, it is possible for the proposition to be assigned a non-zero probability according to another knowledge source $C'$.

Now we can see that *only certain types of implicit conditions can be represented as explicit conditions*. It follows that *if some knowledge isn't available when the prior probability is determined, it is impossible to be put into the system through conditionalization*.

In fact, when $S$ is finite, a Bayesian system can only accept a finite amount of (different) new knowledge after its prior probability distribution is determined, if the above mentioned usage of Bayes' theorem is the only rule used by the system to process new knowledge. To see it, we only need to remember that the new knowledge must be in $S$, and each time a proposition $A$ is provided to the system as a piece of new knowledge, at least $A$ and $\neg A$ (as well as $\neg A \wedge B$, and so on) cannot be used as new knowledge in the future. As a result, the number of different new knowledge that the system can learn is less than $|S|/2$.

Moreover, if we insist that all implicit conditions must satisfy the three preconditions, the prior probability distribution will degenerate into a consistent assignment of 0 or 1 to each proposition in $S$, and, after the assignment, the system will be unable to accept any new knowledge at all.

From a practical point of view, the restrictions set by the three preconditions are not trivial, since they mean that although the background knowledge can be probabilistic-valued, all new knowledge must be binary-valued; no novel concept and proposition can appear in new knowledge; and if a proposition is given a probability 1 or 0, such a belief cannot be changed in the future, no matter what happens. We could build such a system, but unfortunately it would be a far cry from the everyday reasoning process of a human being.

Bayes' theorem can be used as a revision rule, but with very strong restrictions. These limitations are so obvious and well-known that they seems trivial and are often ignored.

Some people claim that the Bayesian approach is sufficient for reasoning with uncertainty, and many people treat Bayes' theorem as a generally applicable revision rule, because explicit conditions and implicit conditions of a probability assignment are seldom clearly distinguished in the discussions (Cheeseman 1985, 1986, and 1988; Pearl 1986, 1987, and 1988), where it is very common that

1. the prior probability of proposition $H$ is formulated as $P(A|K)$,
2. conditional probability is formulated as $P(A|E,K)$,
3. belief revision is described as the process by which $P(A|B)$ and $P(A|C)$ are combined to produce $P(A|B,C)$.

What does the $K$ (and $B$, $C$) mean in these formulas? Pearl interpreted $P(A|K)$ as "a person's subjective belief in $A$ given a body of knowledge $K$" (Pearl 1987). Cheeseman said that conditional probability statements contain the context (conditions) associated with their values, which "make explicit our prior knowledge" (Cheeseman 1986). If $K$ really means implicit condition, then it should not be written in such a form, which suggests that it is a proposition in $S$; if it means explicit condition, then the revision process is not correctly described in the above formula, where $P(A|B)$ and $P(A|C)$ share the same implicit condition (so it can be omitted).

Without a clear distinction between implicit conditions and explicit conditions, the illusion arises that all the knowledge supporting a probability distribution can be represented by explicit conditions, and can therefore be learned by the system using Bayes' theorem. As a result, the capacity of Bayes' theorem is overestimated.

## 4 UPDATING VS. REVISION

Now let us examine some other tools in probability theory that have been used for revision, to see whether we can avoid the three preconditions.

After a prior probability distribution $P_G$ is assigned to a proposition space $S$, if some new evidence shows that "The probability of a proposition $A$ ($A \in S$) should be changed to $m$", assuming the conditional probabilities that with $A$ or $\neg A$ as explicit condition are unchanged, we can update the probability assignment for every proposition $x$ in $S$ to get a new distribution function by using Jeffrey's rule (Kyburg 1987 and Pearl 1988):

$$P_{G'}(x) = P_G(x|A) \times m + P_G(x|\neg A) \times (1-m) \quad (7)$$

If we interpret "$A$ happens" as "$A$'s probability should be changed to 1", then Bayes' theorem, when used as a revision rule, becomes a special case of Jeffrey's rule, where $m = 1$.

As a generalized version, Jeffrey's rule avoids the first precondition of Bayes' theorem, that is, the new evidence must be a binary proposition. However, the other limitations are still applicable, that is, $A \in S$ and $P_G(A) > 0$, otherwise $P_G(x|A)$ is undefined.

More than that, the rule is an *updating* rule, by which I mean a very special way of changing a system's beliefs. In an updating, when the new knowledge "the probability of $A$ should be $m$" arrives, the system's opinion on $A$ is completely dominated by the new knowledge, regardless of $P_G(A)$, the previous opinion about $A$ (Dubois and Prade 1991), and then the distribution function is modified accordingly. Such a complete updating seldom happens in human reasoning. For *revision* in general, new evidence usually causes an *adjustment*, rather than an *abandonment*, of the previous opinion.

A related method was suggested to process "uncertain evidence" $E[m]$ ($m \in (0,1)$), where a "virtual proposition" $V$ is introduced to represent the new knowledge as "a (unspecified) proposition $V$ is true, and $P(E|V) = m$" (Cheeseman 1986 and Nilsson 1986). Then a new conditional probability distribution can be calculated (after considering the new knowledge) for each proposition $x \in S$ in the following way:

$$P(x|V) = P(x|E \wedge V) \times P(E|V) + P(x|\neg E \wedge V) \times P(\neg E|V) \quad (8)$$

Under the assumption that

$$P(x|E \wedge V) = P(x|E) \quad (9)$$

and

$$P(x|\neg E \wedge V) = P(x|\neg E) \quad (10)$$

equation (8) can be simplified into

$$P(x|V) = P(x|E) \times m + P(x|\neg E) \times (1-m) \quad (11)$$

If we transform the explicit condition into the implicit condition by conditionalization, we end up almost with Jeffrey's rule. The only difference is that the prior probability is not *updated* directly, but is instead *conditionalized* by a virtual condition (the unspecified proposition $V$). However, no matter which procedure is followed and how the process is interpreted, the result is the same.

Some other systems process uncertain evidence by providing *likelihood ratios* of virtual propositions (Pearl 1986 and Heckerman 1988), and this method also leads to conditionalization of a virtual condition, therefore the rule is an updating rule, too.

What I mean isn't that updating is not a valid operation in uncertain reasoning, but that it is different from revision. In certain situations, it is more proper to interpret belief changes as updatings (Dubois and Prade 1991), but revision seems to be a more general and important operation. When there are conflicts among beliefs, it is unusual that one piece of evidence can be *completely* suppressed by another piece of evidence, even though it make sense to assume that new evidence is usually "stronger" than old evidence.

## 5 A DEFECT OF THE BAYESIAN APPROACH

Technically, all the above discussed limitations of Bayes' theorem and Jeffrey's rule are known to the uncertainty reasoning community, but it is also a fact that they are often ignored or misunderstood. As a result, the limitation of Bayesian approach is usually underestimated. One reason for this, as I claimed previously, is the confusing of *propagation* and *revision*, *updating* and *revision*, as well as *explicit condition* and *implicit condition* of a probability assignment. Once again, other authors may prefer to name these concepts differently, and what I want to insist is the necessity of making such distinctions.

By *Bayesian approach*, I mean systems where

1. current knowledge is represented as a (real-valued) probability distribution on a proposition space, and
2. new knowledge is learned by conditionalization.

Using previously introduced formalism, the knowledge base of the system at a given instant can be represented as $P_G(x)$, where $x \in S$.

I claim that the system cannot carry out the general revision task, that is, to learn the new knowledge $A[m]$, which may conflict with the system's current beliefs.

Actually, the conclusion directly follows from the discussions in the previous sections:

1. It cannot be done by directly using Bayes' theorem,

since $m$ may be in $(0,1)$, $A$ may not be in $S$, and $P_C(A)$ may be 0.

2. The task cannot be formulated as "from $P(A|C)$ and $P(A|C')$ to get $P(A|C \wedge C')$", then processed by Bayes' theorem, since it is not always possible (or make sense) to represent implicit conditions as explicit conditions.

3. It cannot be done by using Jeffrey's rule or its variations, since usually we don't want the system's previous opinion $P_C(A)$ (if $A \in S$) to be completely ignored.

If the above arguments are accepted as valid, there are some further questions:

1. Do we really need a system to do revision in the general sense?
2. Why it cannot be done in the Bayesian approach?
3. How to do it in a formal reasoning system?

For the first question, if we want to apply the Bayesian approach to a practical domain, one of the the following requirements must be satisfied:

1. The implicit condition of the initial probability distribution, that is, the domain knowledge used to determine the distribution, can be assumed to be immune from future modifications; or

2. All modifications of the implicit condition can be treated as updating, in the sense that when new knowledge conflict with old knowledge, the latter is completely abandoned.

From artificial intelligence's point of view, such domains are exceptions, rather than general situations. In most cases, we cannot guarantee that all knowledge the system get is unchangeable, or later acquired knowledge is always "truer" than earlier acquired knowledge. More than that, under certain conditions we even cannot guarantee that the system's beliefs are free from internal conflicts (Wang 1993). Therefore, we really hope a formal system can revise its knowledge in the general sense.

For the second question, let's look at the revision operation as defined in the first section. For a proposition $A$, if from some evidence $C_1$ its certainty value is evaluated as $m_1$, but from some other evidence $C_2$ its certainty value is evaluated as $m_2$, then what should be the system's opinion on $A$'s certainty, when both $C_1$ and $C_2$ are taken into consideration? Obviously, the result not only depends on $m_1$ and $m_2$, but also depends on the relation between $C_1$ and $C_2$.

For examples, if $C_2$ is already included in $C_1$, then $m_1$ is the final result; if $C_1$ and $C_2$ come from different sources, and $C_1$ consists of large amount of statistical data, but $C_2$ only consists of a few examples, then the result will be closer to $m_1$ than to $m_2$.

In the Bayesian approach, above $m_i$'s become probability assignments, and $C_i$'s become implicit conditions of these assignments. However, in probability theory, there is no way to backtrack a probability distribution's implicit condition only from the distribution function itself, since it is possible for different implicit conditions to generate the same prior probability distribution. The only information available about implicit conditions is their arrival times: the current distribution is "old", and the coming evidence is "new". In such a case, revisions have to be simplified into updatings.

The Bayesian approach have no available revision rule, because its representation is not sufficiently informative — little information is there about the implicit condition of the current probability distribution.

It is impossible for the third question to be discussed throughoutly in this paper. Here I only want to make one claim: *if the system's beliefs are still represented by a set of propositions, then the uncertainty of each proposition must be indicated by something that is more complicated than a single (real) number.* As discussed above, the information about $C_i$ cannot be derived from $m_i$. To revise a belief, the belief's implicit condition must be somehow represented.

## 6   AN EXAMPLE

There are several paradigms using more than one numbers to represent a proposition's uncertainty, such as Dempster-Shafer theory (Shafer 1976), probability interval (Weichselberger and Pöhlmann 1990), and higher-order probability (Paaß 1991). I'm also working on an intelligent reasoning system myself, which use a pair of real numbers as a proposition's "truth-value". The comparison and evaluation of these systems are beyond the scope of this paper, but I will use an example processed by my system to show concretely the problem of the Bayesian approach that discussed in the previous sections.

The system, "Non-Axiomatic Reasoning System" (or "NARS" for short), is described in a technical report (Wang 1993) in detail. In the following, I'll briefly mention (with necessary simplifications) some of its properties that are most directly related to our current topic.

In NARS, domain knowledge is represented as *judgments*, and each judgment has the following form:

$$S \subset P \quad <f, c>$$

where $S$ is the *subject* of the judgment, and $P$ is the *predicate*. "$S \subset P$" can be intuitively understood as "S are P", which is further interpreted as "$P$ inherits $S$'s instances and $S$ inherits $P$'s properties". "$<f, c>$" is the judgment's *truth value*, where $f$ is the *frequency* of the judgment, and $c$ is the *confidence*.

In the simplist case, the judgment's truth value can be determined like this: if the system has checked the "inheritance relation" between $S$ and $P$ for $n$ times ($n > 0$) by looking at $S$'s instances, and in $m$ times ($n \geq m \geq 0$) the checked instance is also in $P$, then $f = m/n$, and $c = n/(n+k)$.[1]

---
[1] $k$ is a parameter of the system. In the current version of the system, $k = 2$.

Here the meaning of $f$ is obvious: it is the inheritance relation's "success frequency", according to the system's experience. The "confidence" $c$ is introduced mainly for the purpose of revision. Given two propositions that have the same *frequency*, their *confidence*, that is, how difficult the corresponding frequency can be revised by future evidence, may be quite different. When $n$ is large, $c$ is large, too, indicating that the frequency $f$ is based on many samples, therefore will be more stable during a revision than a frequency that is only supported by several samples.[2]

Here is an example that shows how NARS works.

### Stage 1

The system is provided with two judgments by the user:

$J_1$ :   $dove \subset flyer$   $<0.9, 0.9>$

$J_2$ :   $dove \subset bird$   $<1, 0.9>$

which means the user tells the system that "About 90% doves are flyers", and "Doves are birds". The confidence of the judgments are pretty high, indicating that they are strongly supported by background knowledge of the user.

From these judgments, the system can use its induction rule [3] to generate a conclusion

$J_3$ :   $bird \subset flyer$   $<0.9, 0.288>$

That is, "About 90% birds are flyers", with a low confidence, since the estimation of frequency is only based on information about doves.

### Stage 2

The system is provided with other two judgments by the user:

$J_4$ :   $swan \subset flyer$   $<0.9, 0.9>$

$J_5$ :   $swan \subset bird$   $<1, 0.9>$

That is, "About 90% swans are flyers", and "Swans are birds", also with high confidence values.

Again by induction, the system get:

$J_6$ :   $bird \subset flyer$   $<0.9, 0.288>$

$J_6$ and $J_3$ looks identical with each other, but they comes from different sources (NARS can recognize this).

---

[2]This is only the simplest case, but not the general method, to determine the truth value of a proposition.

[3]NARS' induction rule: from two premises

$M \subset P$   $<f_1, c_1>$,    $M \subset S$   $<f_2, c_2>$

the following conclusion can be generated

$S \subset P$   $<f, c>$

where

$f = f_1$
$c = \frac{f_2 c_1 c_2}{f_2 c_1 c_2 + k}$

Since $k = 2$, all inductive conclusions are hypotheses with low confidences ($c < 1/3$).

Now the system can use its revision rule [4] to merge $J_3$ and $J_6$ into

$J_7$ :   $bird \subset flyer$   $<0.9, 0.447>$

Here the frequency is not modified (since the two premises give the same estimation), but the confidence of the conclusion is higher, because now the frequency estimation is supported by more evidence (*dove* and *swan*).

### Stage 3

The system is provided with two more judgments by the user:

$J_8$ :   $penguin \subset flyer$   $<0, 0.9>$

$J_9$ :   $penguin \subset bird$   $<1, 0.9>$

That is, "No penguin is a flyer", and "Penguins are birds", also with high confidence values.

By induction, the system get from them

$J_{10}$ :   $bird \subset flyer$   $<0, 0.288>$

Therefore, penguin provide a negative example for "Birds are flyers", but doesn't completely "falsify" the hypothesis, because the hypothesis is treated by NARS as a *statistical proposition*, rather than an *universal generalization* in Popper's sense (Popper 1968).

Using $J_7$ and $J_{10}$ as premises, the conclusion is further revised:

$J_{11}$ :   $bird \subset flyer$   $<0.6, 0.548>$

The frequency of the result is lower than that of $J_7$ (where only positive examples are available), but higher than $J_{10}$ (where only negative examples are available), and closer to $J_7$ than to $J_{10}$, since the former has a higher confidence (that is, supported by more evidence). The order that the two premises are acquired by the system is irrelevant — "new knowledge" doesn't have a higher priority in revision. The confidence of $J_{11}$ is higher than either of the premises, because in revision the conclusion always summarizes the

---

[4]NARS' revision rule: if the two premises

$S \subset P$   $<f_1, c_1>$,    $S \subset P$   $<f_2, c_2>$

come from different sources, the following conclusion can be generated:

$S \subset P$   $<f, c>$

where

$f = \frac{w_1 f_1 + w_2 f_2}{w_1 + w_2}$
$c = \frac{w_1 + w_2}{w_1 + w_2 + 1}$   $(w_i = \frac{c_i}{1 - c_i}, i = 1, 2)$

If the two premises come from correlative sources, the one that has a higher confidence is chosen as the result. This is the rule in NARS that corresponds to updating as discussed in previous sections.

For a detailed discussion about the rules (as well as other rules in NARS, such as those for deduction and abduction), see the technical report (Wang 1993).

premises, therefore supported by more evidence (compared with the premises), no matter whether the premises are consistent with each other (as when $J_7$ is generated) or in conflict with each other (as when $J_{11}$ is generated).

Without detailed discussing about how the truth values are calculated in NARS, we can still get some general impressions about how revisions are carried out in NARS:

1. Revision is used to summarize information about the same proposition that comes from different sources.
2. All propositions are revisable.
3. When two propositions are summarized, the frequency of the result looks like a weighted sum of the frequencies of the premises, with the weights determined by the confidence of the premises.
4. The confidence of a revision conclusion is always higher than the confidence of either of the premises.
5. Frequency and confidence are two independent measurements, that is, it is impossible to determine one from the other. In the above example, $J_6$ and $J_7$ have the same frequency but different confidence; $J_6$ and $J_{10}$ have the same confidence but different frequency.
6. Generally, the two measurements of certainty have different functions in representing a system's beliefs: frequency is indicating the extent to which a belief is positive ("Yes, it is the case.") or negative ("No, it is not the case."), and confidence is indicating the extent to which a belief is stable ("Yes, I'm sure.") or fragile ("No, it is only a guess.").

I believe that these properties are also shared by actual human uncertain reasoning.

In probability theory, especially in the Bayesian approach, the two factors ($f$ and $c$) are somehow summarized into a single "probability distribution". When a proposition is assigned a probability closing to 1, usually it means that almost all the background knowledge support the prediction that the proposition is true, and the system already know a lot about the proposition. When a proposition is assigned a probability closing to 0.5, however, there are different possibilities: sometimes it means that the system knows little about the proposition; sometimes it means the system knows a lot, but the positive evidence and the negative evidence are almost equally strong.

Combining these two factors into a single measurement seems fine (and sometimes even more convenient) for the comparison and propagation operation, but it doesn't work well for the revision operation, as discussed above.

Even this conclusion is not completely new. All paradigms that use more than one numbers to represent uncertainty come from the observation that "Ignorance cannot be properly represented by a real-value probability distribution". However, this observation is also often misinterpreted.

To argue against the opinion that "more than one number is needed to represent uncertainty", Cheeseman claimed (Cheeseman 1985) that a point value and a density function will give the same result in decision making, which I agree to certain extent. However, I believe he was wrong by saying that standard deviation can be used to capture "the change of expectations" (or revision, as defined in this paper). If we test a proposition $n$ times, and the results are the same, then the standard deviation of the results is 0, that is, independent to $n$. But our confidence about "the result will remain the same" will obviously increase with $n$. Actually, what the standard deviation measures is the *variations* among the samples (which has little to do with revision), but what the confidence measures, intuitively speaking, is the *amount* of the samples.

Pearl said our confidence in the assessment of $BEL(E)$ is measured by the (narrowness of the) distribution of $BEL(E|c)$ as $c$ ranges over all combinations of contingencies, and each combination $c$ is weighted by its current belief $BEL(c)$ (Pearl 1988). I agree with him that ignorance is the lack of confidence, and confidence can be measured by how much a belief assignment can be modified by possible future evidence. However, in his definition, he still assumes that all relevant future evidence causing a belief change can be represented as an *explicit condition*, and can be processed through conditionalization. As a result, his measurement of confidence cannot captures the ignorance about *implicit conditions*.

No matter whether other paradigms can solve the problem, I claim that when the "ignorance" to be represented is about an implicit condition, it cannot be handled properly by the Bayesian approach. For a specific domain, if revision is a crucial operation for the solving of the practical problems, the Bayesian approach cannot be used, and other paradigms should be considered.

## 7 SUMMARY

The following conclusions have been drawn:

1. Propagation and revision are different operations in reasoning systems where uncertainty is represented and processed; the former generates new beliefs (with truth values), and the latter modifies truth values of previous beliefs.
2. The explicit condition and the implicit condition of a probability assignment are different, and the latter has a much greater capacity for representing knowledge.
3. When used as revision rules, Bayes' theorem merges explicit condition with implicit condition. Its ability is therefore limited.
4. Jeffrey's rule (and its variations) is an updating rule (replacing old knowledge by new knowledge), rather than a revision rule in the general sense (combining knowledge from different sources).
5. In the Bayesian approach, there is no way to do revision, because the "frequency" factor and the "confidence" factor in a probability distribution cannot be distinguished from each other, and these two factors have different functions in revision.

6. For a system to solve the revision problem, it is not sufficiently informative to represent a proposition's uncertainty by a single number.

### Acknowledgements

Thanks to Jim Marshall, David Chalmers, Steve Larson, and the anonymous referees for helpful comments and suggestions.